\documentclass[manuscript,screen, authorversion]{acmart}

\AtBeginDocument{%
  \providecommand\BibTeX{{%
    \normalfont B\kern-0.5em{\scshape i\kern-0.25em b}\kern-0.8em\TeX}}}

\setcopyright{acmcopyright}
\copyrightyear{2023}
\acmYear{2023}
\acmDOI{XXXXXXX.XXXXXXX}

\acmConference[Journal]{Journal}
\acmISBN{978-1-4503-XXXX-X/18/06}

\usepackage{bbding}
\usepackage{xcolor}
\usepackage{pifont}

\newcommand{\cmark}{\ding{51}}%
\newcommand{\xmark}{\ding{55}}%

\begin{document}

\title{Multimodality Representation Learning: A Survey on Evolution, Pretraining and Its Applications}

\author{Muhammad Arslan Manzoor}
\affiliation{%
  \institution{Mohamed bin Zayed University of Artificial Intelligence}
  \city{Masdar}
  \state{Abu Dhabi}
  \country{UAE}
}

\author{Sarah AlBarri}
\affiliation{%
  \institution{Mohamed bin Zayed University of Artificial Intelligence}
  \city{Masdar}
  \state{Abu Dhabi}
  \country{UAE}}
  
\author{Ziting Xian}
\affiliation{%
  \institution{Sun Yat-sen University}
  \city{Guangzhou}
  \country{China}}

\author{Zaiqiao Meng}
\affiliation{%
  \institution{University of Glasgow}
  \city{Glasgow}
  \country{UK}}

\author{Preslav Nakov}
\affiliation{%
  \institution{Mohamed bin Zayed University of Artificial Intelligence}
  \city{Masdar}
  \state{Abu Dhabi}
  \country{UAE}}

\author{Shangsong Liang}
\authornote{Shangsong Liang is the corresponding author of the paper.}
\affiliation{%
  \institution{Mohamed bin Zayed University of Artificial Intelligence}
  \city{Masdar}
  \state{Abu Dhabi}
  \country{UAE}}

\renewcommand{\shortauthors}{Manzoor et al.}

\begin{abstract}

Multimodality Representation Learning, as a technique of learning to embed information from different modalities and their correlations, has achieved remarkable success on a variety of applications, such as Visual Question Answering (VQA), Natural Language for Visual Reasoning (NLVR), and Vision Language Retrieval (VLR). Among these applications, cross-modal interaction and complementary information from different modalities are crucial for advanced models to perform any multimodal task, e.g., understand, recognize, retrieve, or generate optimally. Researchers have proposed diverse methods to address these tasks. The different variants of transformer-based architectures performed extraordinarily on multiple modalities. This survey presents the comprehensive literature on the evolution and enhancement of deep learning multimodal architectures to deal with textual, visual and audio features for diverse cross-modal and modern multimodal tasks. This study summarizes the (\emph{i})~recent task-specific deep learning methodologies, (\emph{ii})~the pretraining types and multimodal pretraining objectives, (\emph{iii})~from state-of-the-art pretrained multimodal approaches to unifying architectures, and (\emph{iv})~multimodal task categories and possible future improvements that can be devised for better multimodal learning. Moreover, we prepare a dataset section for new researchers that covers most of the benchmarks for pretraining and finetuning. Finally, major challenges, gaps, and potential research topics are explored. A constantly-updated paperlist related to our survey is maintained at \url{https://github.com/marslanm/multimodality-representation-learning}. 
\end{abstract}


\maketitle

\section{Introduction}
Multimodal systems utilize two or more input modalities, such as audio, text, images, or video to produce an output modality, which could be different from the inputs. Cross-modal systems, a subpart of multimodal systems, utilize information from one modality to enhance the performance in the other modality. For example, a multimodal system would use image and textual modalities to assess a situation and perform a task, whereas a cross-modal system would use an image modality to output a textual modality \cite{recent_paper,baltruvsaitis2018multimodal}. Audio-Visual Speech Recognition (AVSR) \cite{shi2022learning}, detecting propaganda in memes \cite{dimitrov2021detecting}, and Visual Question Answering (VQA) \cite{VQA2}  are examples of the multimodal systems. Multimodal representation learning techniques reduce the heterogeneity gap between different modalities by processing raw heterogeneous data hierarchically \cite{guo2019deep}. Heterogeneous features from different modalities offer additional semantics in the form of contextual information \cite{guo2019deep}. Thus, complementary information can be learned through multiple modalities. For example, the visual modality can help speech recognition by providing lip motion \cite{bayoudh2021survey} in the AVSR. Recent advanced variants of deep learning approaches have addressed classical multimodal challenges (correlation, translation, alignment, fusion) by mapping different modalities into a standard representation space.

\begin{figure*}[t]
    \centering
    \includegraphics[trim=0.75in 1.1in 1.6in 1.8in,clip,width=6.1in, height= 1.6in]{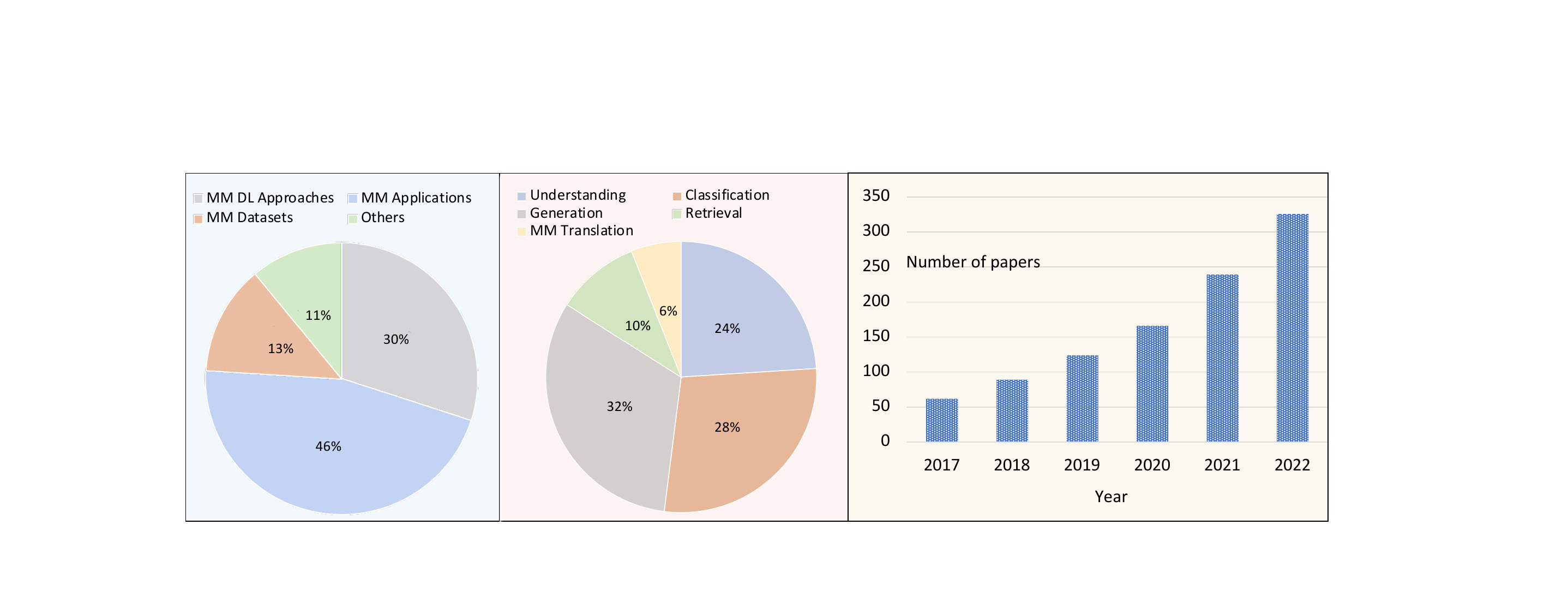}
    \caption{The pie chart on the left represents the percentage of papers for each section included in this survey. The pie chart in the center represents the percentage of papers for each multimodal application. The rightmost figure expresses the growth of deep learning-based multimodal papers in the last six years on Google Scholar.}
    \label{fig:visualization}
    \vspace{-1em}
\end{figure*}

In the past, many task-specific deep learning strategies have significantly enhanced the performance for a variety of multimodal tasks  \cite{li2019visualbert}. More recently, pre-training and fine-tuning methods for Natural Language Processing (NLP) and Computer Vision (CV) have garnered significant attention, primarily due to their semantically rich representation and the availability of large-scale public models \cite{lu2019vilbert}. This study summarized the pipeline of the task-specific methods for multimodal representation. We discuss pretraining types and pretext task required to make the pretrained model robust on the variety of multimodal or cross-modal downstream tasks. We demonstrate that a majority of pretraining methods leverage transformers, leading to the inception of unifying architectures. These models, capable of handling all modalities across diverse downstream tasks, mitigate the need for individual task-specific fine-tuning, thereby reducing both computational complexity and processing time \cite{lu202012}. Additionally, evaluation on downstream tasks, multimodal benchmarks, extensive range of multimodal applications, including NLP tasks that are enriched by the incorporation of visual and audio modalities, e.g.,~sentiment analysis, document understanding, fake news detection, retrieval, translation, and other reasoning applications are comprehensively presented in this survey. 
Figure~\ref{fig:visualization} exhibits the categorical percentage of deep learning multimodal papers included in this survey. The bar graph shows the development and the availability of deep learning multimodal approaches on Google Scholar yearly.\\

\vspace{-1.4em}
\section{Background - The Advance evolution of multimodal architectures}\label{sec:background}

The concept of multimodal research was initially inspired by the field of audio-visual speech recognition (AVSR) \cite{mcgurk1976hearing}, where well-aligned integration of vision and hearing was a fundamental requirement. Encouraging outcomes from AVSR studies prompted the research community to broaden the application of this approach to other multimodal tasks e.g., content indexing and  retrieval \cite{atrey2010multimodal}, video summarization \cite{evangelopoulos2013multimodal} and shot-boundary detection \cite{lienhart1998comparison}. 
In the early 2000s, the AMI Meeting \cite{carletta2006ami} and SEMAINE \cite{mckeown2010semaine} corpora were generated with the intention of understanding human multimodal behaviors and the dynamics of interpersonal interactions. These foundational datasets not only led to the inception of the Audio-Visual Emotion Challenge (AVEC) \cite{schuller2011avec}, but also later propelled its expansion into the realm of automatic detection of depression and anxiety \cite{valstar2014avec}. Over time, multimodal methodologies have advanced, enabling simultaneous processing of diverse modalities such as speech, vision, and text \cite{baltruvsaitis2018multimodal,chen2022revisiting}. 
The latest category encompasses data derived from multimedia sources that extend beyond the complexity of two or three modalities. These complex multimodal approaches motivate researchers to tackle associated challenges through more effective strategies such as pretraining and fine-tuning. These methods have been found to deliver superior performance compared to traditional task-specific frameworks. Advanced approaches are particularly adept at tackling complex multimodal tasks, exhibiting superior performance to previous models.


The primary obstacle for multimodal methods lies in understanding the relationships between various encoding schemes of multiple modalities, a problem commonly identified as the heterogeneity gap in learning representation \cite{rasiwasia2010new}. The fusion module in multimodal learning systems aids in reducing the heterogeneity gap by correlating the similar semantics across different modalities, a strategy that has been proven to boost performance in a majority of tasks \cite{habibian2016video2vec}. Recent research has demonstrated that deep learning methods effectively understand and process representations due to their robust learning capabilities \cite{lecun2015deep}. One of the key advantages of deep learning is its ability to hierarchically learn representations as general-purpose learning without needing additional structural information. These factors contribute to making deep learning an extremely compatible approach for multimodal representation learning.

\begin{figure*}[t!]
\centering
\includegraphics[width=14.2cm,height=9.9cm]{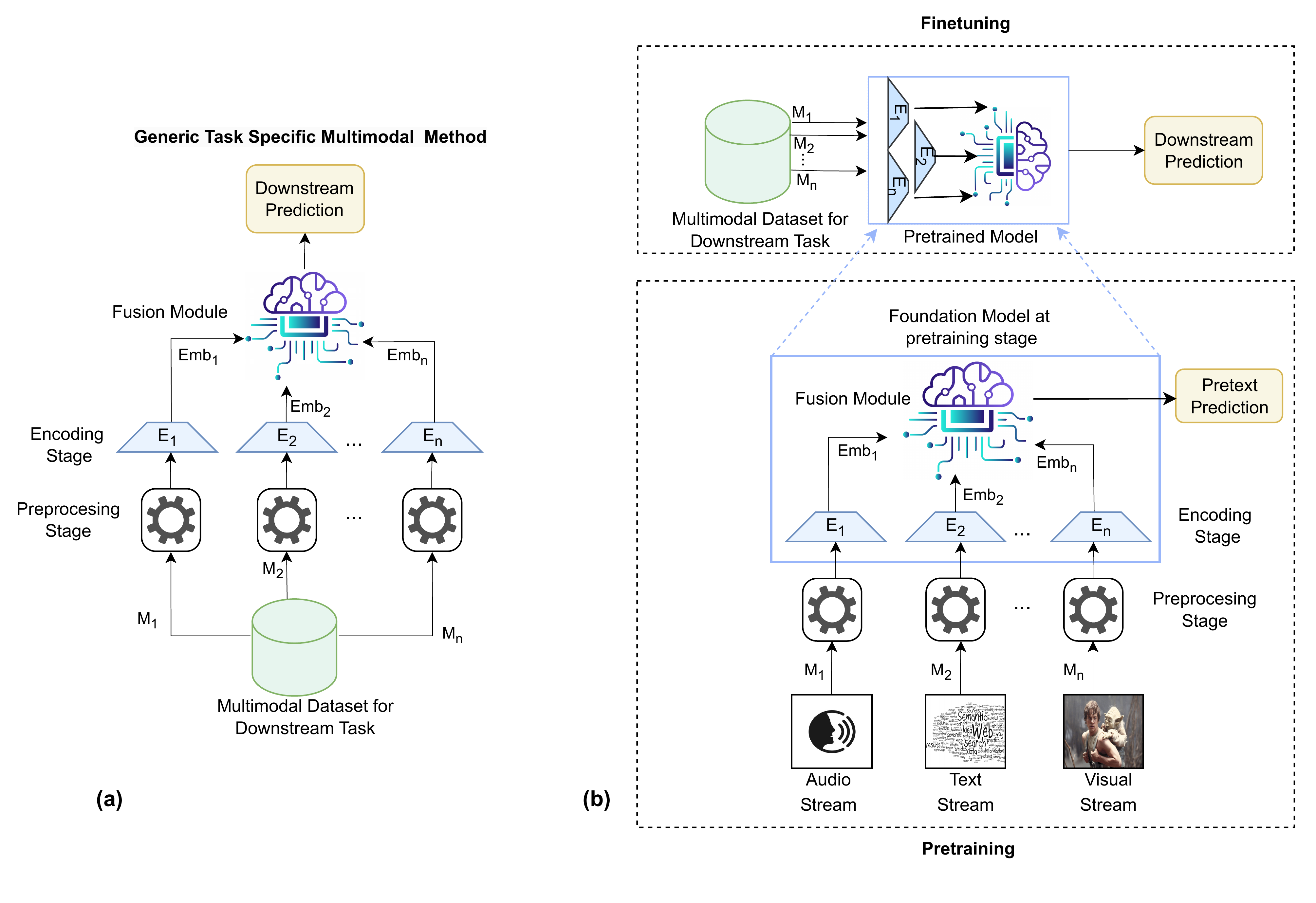}
  \caption{\textbf{(a)} Task-specific multimodal methods are trained on benchmarks created for specialized downstream tasks. After preprocessing, each encoder ($\mathrm{E}_i$) receives one modality ($\mathrm{M}_i$) data to produce an embedding. The fusion module is responsible for the interaction of features from different modalities and is trained to predict the downstream task. \textbf{(b)} In pretraining, the raw data from the web or dataset, in the form of any modality ($\mathrm{M}_i$), is preprocessed and passed to a specialized encoder ($\mathrm{M}_i$) at the encoding stage. Encoders produce embeddings which are integrated by the fusion module to produce a meaningful unified representation by predicting pretext tasks. The blue-bounded box in the pretraining stage represents the foundation model, which is readily available at the fine-tuning stage as a pretrained model. A multimodal dataset for the downstream task can leverage this generic pretrained model for downstream prediction.}
  \vspace{-1em} 
  \label{fig:represent}
\end{figure*} 

The early task-specific multimodal methods were trained to address only one multimodal task. Figure~\ref{fig:represent}\textbf{(a)} broadly depicts the pipeline of task-specific multimodal methods. For example, in VQA, a visual encoder, usually based on some variant of a Convolutional Neural Network (CNN) \cite{simonyan2014very, he2016deep}, is used to extract visual features. Meanwhile, textual features are encoded by text encoding blocks that typically build on Bag-of-Words \cite{antol2015vqa}, seq2seq models \cite{hochreiter1997long}, possibly with attention mechanism \cite{vaswani2017attention}. 
Subsequently, the encoded representations are channeled to Multimodal Fusion (explored extensively for multimodal learning), which fuses and builds interactions between features of different modalities to produce a unified representation \cite{gan2022vision}. These Multimodal Fusion methods have evolved as (\emph{i})~\emph{simple fusion}, which fuses features through concatenation or element-wise sum or product \cite{jabri2016revisiting}, (\emph{ii})~\emph{inter-modality attention}, which aligns features of different modalities and constructs more informative joint representations \cite{nguyen2018improved}, (\emph{iii})~\emph{intra-modality attention}, which develops relational reasoning by forming a graphical representation within modalities \cite{hu2019language}, (\emph{iv})~\emph{transformers}, which leverage the aforementioned techniques by concurrently focusing on other modalities and associated regions within the same modality \cite{gao2019dynamic}. Since transformers learn and establish intricate interactions between different modalities \cite{wang2021vlmo,kim2021vilt}, they are viewed as the structural foundational block of universal multimodal pretraining methods.

With the availability of large-scale datasets and computing resources, the pretraining paradigm has yielded efficient performance in various domains, including text \cite{devlin2018bert}, vision \cite{krizhevsky2017imagenet}, and speech \cite{baevski2020wav2vec}. This has spurred interest in multimodal pretraining, allowing for the processing of raw data (both labeled and unlabeled) using semi-supervised, self-supervised, or unsupervised approaches. As depicted in Figure~\ref{fig:represent}\textbf{(b)}, each modality is processed by a dedicated encoder to generate embeddings. These embeddings are then passed to a fusion module specifically trained for multimodal pretext tasks. The generation of contextually rich embeddings through dedicated encoders allows the fusion module to learn more robust associations between different modalities. This, in turn, can lead to more accurate performance in downstream tasks, as the model becomes adept at handling complex multimodal data. In the fine-tuning phase of multimodal learning, the pre-trained model, serving as a flexible backbone, is further trained using a specific multimodal dataset for a target task. Through fine-tuning, the model adjusts its learned representations to better accommodate the relationships and interactions between different modalities present in the task-specific dataset. Consequently, the rich, general-purpose knowledge from pretraining is transformed into task-specific insights, improving the model's performance on the downstream task. The Figure \ref{fig:represent} clearly illustrates two pivotal approaches in multimodal learning: task-specific methods and the pretraining-fine-tuning paradigm. They are primarily distinguished by the type of input data: task-specific methods leverage specialized benchmarks, while pretraining can employ any kind of data. Task-specific methods are trained directly for the downstream task, whereas pretraining is for a pretext task and yields a generic pretrained model that can be adapted to a variety of downstream tasks, thus saving computational time and complexity associated with individually training each model.

\vspace{-0.3em}
\subsection{Comparison with previous surveys}
\vspace{-0.3em}
Numerous surveys have extensively investigated multimodality, with a specific focus on multimodal learning techniques, pretraining technologies, model-specific topologies, and task-specific applications. Representation, correlation, fusion, translation and co-learning are essential components of multimodal learning that are surveyed in \cite{guo2019deep,baltruvsaitis2018multimodal, bayoudh2021survey, co-learn, fusion}. In \cite{baltruvsaitis2018multimodal}, the taxonomy of multimodal learning components is established. Additionally, \cite{bayoudh2021survey} contributes by offering a perspective on these components from the domain of computer vision. The work of \cite{guo2019deep} and \cite{fusion} solely examined representation and fusion, respectively, presenting their mathematical frameworks, architectures, challenges, and prospects. Despite the importance of multimodal learning components in understanding and processing different modalities, these surveys are focused and fail to present a holistic depiction of multimodality.

Vision-language pretraining objectives, strategies, architectures and datasets are extensively studied in \cite{vlp-survey, du2022survey, long2022vision, gan2022vision}.  The multimodal pretraining survey \cite{du2022survey} focused on image-text tasks as it reviews methodologies of encoding of raw image and text, and architectures for the interactions. While downstream tasks are addressed in all pretraining surveys, the level of detail regarding state-of-the-art frameworks is limited, and there is insufficient coverage of diverse applications within the multimodal domain. The studies \cite{ramachandram2017deep}, \cite{recent_paper} and \cite{trendsVisionLang} inspected multimodal deep learning models, which are presented with respect to modalities involved in \cite{recent_paper} and to prominent tasks in \cite{trendsVisionLang}. The work of \cite{ramachandram2017deep} presented methodologies for finding optimal multimodal architectures and cross-modality regularization. Various works assessed multimodal learning from a specific application perspective, such as visual question answering \cite{VQA}, biomedical applications \cite{kalyan2021ammus}, event detection \cite{xiao2022survey}, sentiment analysis \cite{stappen2021multimodal,chandrasekaran2021multimodal}, and object detection \cite{9285213}.

\begin{table}[!t]
\caption{Survey comparison comparing between different survey work based on their content. The content studied is presence or absence of  pretraining concepts, multimodal applications, NLP focused (textual and audio forms), evolutionary perspective, unifying architectures, benchmark datasets and active repository} \vspace{-0.5em}
\label{survey-compare}
\scalebox{0.85}{
\begin{tabular}{ccccccccc}
\hline
Ref & Year & Pretraining & Applications & NLP focused & Evolution & Unifying Arch. & Datasets & Active repo \\ \hline
\cite{ramachandram2017deep} & 2017 & \xmark & \cmark & \xmark & \xmark & \xmark & \cmark & \xmark \\ \hline
\cite{baltruvsaitis2018multimodal} & 2018 & \xmark & \cmark & \cmark & \cmark & \cmark & \xmark & \xmark \\ \hline
\cite{guo2019deep} & 2019 & \xmark & \xmark & \xmark & \cmark & \cmark & \xmark & \xmark \\ \hline
\cite{recent_paper} & 2021 & \xmark & \cmark & \cmark & \cmark & \xmark & \cmark & \xmark \\ \hline
\cite{trendsVisionLang} & 2021 & \cmark  & \cmark  & \cmark  & \xmark & \xmark & \cmark  & \xmark \\ \hline
\cite{du2022survey} & 2022 & \cmark & \cmark & \xmark & \xmark & \cmark & \cmark & \xmark \\ \hline
\cite{bayoudh2021survey} & 2022 & \xmark & \cmark & \cmark & \xmark & \cmark & \cmark & \xmark \\ \hline
\cite{long2022vision} & 2022 & \cmark & \cmark  &\cmark   &  \xmark & \xmark & \cmark & \xmark \\ \hline
\cite{gan2022vision} & 2022 & \cmark & \cmark  &\xmark   &  \xmark & \cmark & \cmark & \xmark \\ \hline
 \cite{vlp-survey} & 2023 & \cmark & \cmark & \xmark & \xmark & \xmark & \cmark & \xmark \\ \hline
 Our Work & 2023 & \cmark & \cmark & \cmark & \cmark & \cmark & \cmark & \cmark \\ \hline
\end{tabular}
}
\vspace{-0.5em}
\end{table}

Our work expands on existing research, providing a more comprehensive exploration of task-specific methods and multimodal pretraining developments. We elaborate on the significance of pretraining tasks for large-scale pretrained multimodal models and their robustness across diverse downstream tasks. Additionally, we catalogue pretraining benchmark datasets, covering their sizes, included modalities, and targeted tasks. This survey also investigates the advancements of unifying architectures and presents a comparative analysis of various SOTA models. We extend our exploration to a wide range of applications, including vision, language, audio, healthcare, and NLP-specific tasks. Table~\ref{survey-compare} provides a comparative view of our work with other surveys, highlighting the inclusion of pretraining concepts, downstream applications, NLP applications, evolutionary perspectives, unifying architectures, datasets, and active repository availability. This broad approach facilitates a holistic view of multimodal learning, guiding future research by shedding light on efficient multimodal handling techniques.

\section{Multimodal Deep Learning Approaches}\label{sec3}
This section explores multimodal architectures, divided into task-specific and Pretraining-Finetuning architectures (Fig \ref{fig:represent}). Subsection \ref{sec3.1} reviews task-specific methods, recently evolving into large-scale pretrained approaches. Subsection \ref{sec3.2} outlines the pretraining process, including pretext tasks and state-of-the-art (SOTA) frameworks for multimodal tasks. Subsection \ref{sec3.3} compares results from SOTA approaches, while \ref{sec3.0} lists the acronyms used.

\vspace{-0.3em}
\subsection{Multimodal Task-Specific Methods} \label{sec3.1}
Multimodal representation learning improves the robustness of deep learning (DL) models as complementary features are present. 
This section describes task-specific multimodal methods that can be generally classified into encoder-decoder-based, attention-based, and reinforcement learning-based models.   
\vspace{-0.3em}
\subsubsection{Encoder-Decoder Based Models}
Encoder-decoder models, such as cascaded CNN-RNN and RNN-RNN, transform input data to output via semantically rich latent representations. Hu and Wu \cite{CRNN} proposed a Cascaded Recurrent Neural Network (CRNN) to learn image-text interactions. This uses a VGG16 network as the encoder and a Stacked Gated Recurrent Unit (SGRU) as the decoder. Ji et al. \cite{GET} suggested a Globally Enhanced Transformation (GET) network, employing Faster-RCNN and ResNet-101 as the encoder. The GET network outperformed the CRNN on the MS-COCO dataset \cite{MS-COCO} with a higher BLEU score.

Autoencoders \cite{autoenc}, a form of encoder-decoder model using unsupervised learning, construct latent representations to reconstruct original data points \cite{autoenc}. In \cite{autoenc_app}, a multimodal architecture is built on autoencoders for processing audio and video modalities, feeding into a conventional encoder-decoder architecture.

\vspace{-0.3em}
\subsubsection{Graph Based Models} 
In multimodal learning, the fusion module, which facilitates the interplay of features across various modalities, has emerged as a critical and central component \cite{zhu2022multi}. 
Task-specific methods extensively employ graph-based approaches for fusion tasks \cite{gan2022vision}. Graph Neural Networks (GNNs) demonstrate exceptional robustness when handling multimodal data, skillfully compensating for incomplete or noisy inputs in one mode by leveraging information from other modes \cite{li2015gated}. In multimodal tasks, GNNs are capable of encapsulating both local and global context through iterative neighborhood aggregation \cite{velivckovic2017graph}. Graph-based multimodal fusion encoder \cite{yin2020novel} for Neural Machine Translation (NMT) performed superior on Multi30K datasets. The method constructs graphs containing two types of nodes: all words in the sentence are included as textual nodes and the nouns in the sentence, identified by the Stanford Parser, are detected by the visual grounding toolkit as visual nodes. Additionally, two types of edges are employed: intra-modal edges to connect nodes in the same modality, and inter-modal edges to connect corresponding nodes in different modalities, capturing the semantic relationship of multi-modal data. These graphs are processed through multiple graph-based multimodal fusion layers, which iteratively facilitate semantic interactions and learn node representations, ultimately producing an attention-based context vector for the decoder. Likewise, Gao et al. \cite{gao2020multi} addressed the scene textual ambiguity in better way than pre-trained word-embeddings by designing Multimodal Graph Neural Network (MM-GNN). VQA-GNN \cite{wang2022vqa} integrates image-level information and conceptual knowledge into a unified multimodal semantic graph for joint reasoning. The effectiveness of this model is demonstrated by top-ranking performances in the VCR task, outperforming previous models, and notably, it also provides a pioneering cross-domain interpretability for visual and textual knowledge in the visual question answering task.
\vspace{-0.3em}
\subsubsection{Attention Based Models} 
The attention mechanism improves model learning by inspecting information flow, features, and resources. These models help the network to deal with long-time dependencies by allocating attention to vital information and filtering all unnecessary or irrelevant stimuli. A recent work by Jiang et al. \cite{MGAN} employed the attention notion through Attention Weight Gate (AWG) module and Self-Gated (SG) module to the Multi Gate Attention Network (MGAN) pipeline. Moreover, MGAN extracted and utilized intra-object relation in the network \cite{MGAN}. Another attention based variant is explored in attention-guided Multimodal Correlation (AMC) \cite{AMC}. The attention mechanism is applied to the modality rather than a semantic context in a vector, varying the importance of each modality depending on the required query from the system. The application of AMC has applied search logs, where the user inputs an image and text to be searched, and the system determines which modality is more valuable.     
\vspace{-0.3em}
\subsubsection{Reinforcement Learning-Based Models}
Reinforcement learning models learn through trial-and-error interactions between an environment-aware agent and its environment, balancing between exploiting prior knowledge and exploring new actions. The algorithm "reinforces" actions yielding high rewards. Unlike supervised learning, it doesn't need labeled data. This concept is integrated into deep learning in \cite{DL-RL}, using a Hierarchical-based Reinforcement Learning (HCL) framework with roles like a manager, worker, and critic. The manager sets goals for the worker, who performs actions to achieve them, while the critic evaluates goal completion. The HCL outperforms baseline models in transforming video modality into text.

As the complexity of multimodal frameworks increases, further deep learning architectures are modified and enhanced, such as Generative Adversarial Networks (GAN) and Probabilistic Graphical Models (PGM) \cite{fang2021gaussian, chen2022multi}. However, the approaches mentioned in this section pose challenges in terms of time and computational resource efficiency due to the need for retraining for each specific task. Recognizing these challenges, many researchers have now shifted their focus to pretraining frameworks. These aim to address and overcome the limitations inherent in task-specific methods, reducing the need for repetitive retraining for different tasks, and thereby presenting a more efficient avenue for future multimodal representation learning. 

\vspace{-0.2em}
\subsection{Multimodal Pretraining Methods}\label{sec3.2}
This section covers a well-detailed discussion of the pretraining framework, with a special emphasis on multimodal pretraining, including pretraining types, pretext tasks, and state-of-the-art methods. Earlier,
Self-Supervised Learning (SSL) boosted pretraining for language tasks \cite{vaswani2017attention} which was later used for vision tasks and produced efficient results \cite{zhang2021vinvl}. This concept was then extended to multimodal scenarios. Pretraining on massive labeled or unlabeled datasets and finetuning on task-specific datasets has become a modern paradigm for various domains \cite{lin2021m6}.  
\vspace{-0.4em}
\subsubsection{Types of Pretraining}
Kalyan et al. \cite{kalyan2021ammus} proposed different types of pretraining approaches used by researchers to design and train the model on a large scale. These approaches include: (1) Pretraining From Scratch (PTS) in which massive unlabeled text is used to train the model from scratch using random initialization for all the layers of the language model, e.g., ELECTRA \cite{clark2020electra}, RoBERTa \cite{liu2019roberta} and BERT\cite{devlin2018bert}. (2) Continual Pretraining (CPT), models are initialized using existing pretrained models. This approach reduces the required computational resources. The domain-specific task uses this method to alleviate the challenges of short target vocabulary. BioBERT \cite{lee2020biobert}, HateBERT \cite{caselli2020hatebert} and infoXLM \cite{chi2020infoxlm} are the examples of CPT. (3) Simultaneous Pretraining (SPT) used in \cite{wada2020pre}. It consumes domain-specific and general text simultaneously while training from scratch. (4) Task Adaptive Pretraining (TAPT) used in \cite{gururangan2020don} demands a small amount of data. (5) Knowledge Inherited Pretraining (KIPT)  \cite{qin2021knowledge} uses the Knowledge distillation technique.
\vspace{-0.4em}
\subsubsection{Pretraining Tasks}
In the pretraining phase, predefined or pretext tasks are solved to learn language representation. These tasks are based on self-supervised learning and challenging enough to make the model robust by exploiting the training signal, which include: (1) Casual Language Modeling (CLM) is unidirectional and uses context to predict the next word employed by GPT-1 \cite{radford2018improving} for the first time. (2) Masked Language Model (MLM) enhanced the representation by considering the context in both directions and consuming only 15\% of the tokens applied by \cite{devlin2018bert} for the first time. (3) Replaced Token Detection (RTD) mitigates the pretraining challenges of MLM (less supervisory signal) by classifying the replaced token as original or not \cite{clark2020electra}. (4) Shuffled Token Detection (STD) reduces the discrepancy between pretraining and finetuning by exploiting discriminative tasks (detection of shuffled tokens) \cite{panda2021shuffled}.Other pretraining tasks include RTS (Random Token Substitution), NSP (Next Sentence Prediction), SLM (Swapped Language Models), SOP (Sentence Order Prediction), which established as baseline pretext tasks for the multimodal pretraining.\\
\textit{3.2.2.1 Multimodal Pretraining Tasks.}
Multimodal pretraining tasks offer the benefits of enhanced representation learning, cross-modal transfer, improved generalization, effective multimodal fusion, and improved performance in a wide range of multimodal applications. These tasks include: (1) Cross-Modal Masked Language Model (CMML), which is similar to MLM in BERT; moreover, pretraining models also consider contextual visual features and unmasked tokens. It is an effective approach to vision language pretraining by assisting the model for better align and regard the relationship between different modalities \cite{du2022survey}. (2) Cross-Modal Masked Region Prediction (MRP) \cite{zhan2021product1m}: Similar to previous objective, MRP masks RoI with zeros. The contextual visual features and text inference predicts the masks. (3) Image-Text Matching (ITM) or Visual Linguistic Matching (VLM) \cite{lin2020interbert}: This objective considers the coarse-grained relationship between image and text features by matching and aligning. It measures the relevancy between different modalities at high level. (4) Cross-Modal Contrastive Loss (CMCL): This objective learns universal vision and language features by pushing the relevant image-text pairs close and non-relevant apart. The highest similarity score between embeddings of different modalities help make the decision. The similar alternative multimodal pretraining tasks are Masked Object Classification (MOC) \cite{zhan2021product1m}, Phrase-Region Alignment (PRA) \cite{liu2021kd}, Word Region Alignment (WRA) \cite{chen2020uniter}, Video-Subtitle Matching (VSM) \cite{li2020hero}, Mask Frame Modeling (MFM) \cite{li2020hero}, Frame Order Modeling (FOM) \cite{li2020hero}, Visual Translation Language Model (VTLM) \cite{zhou2021uc2}, etc.

Pretrained approaches have been used extensively to perform exceptionally well for NLP downstream tasks. The researchers proposed pretrained models including BERT \cite{devlin2018bert}, ALBERT \cite{lan2019albert}, RoBERTa \cite{liu2019roberta}, and T5 \cite{raffel2020exploring}, to learn general textual representation and to boost downstream tasks' performance.  
Similarly, pretrained approaches have been adopted to reduce the time and computation for vision-based downstream tasks \cite{carion2020end,zhu2020deformable}. Simultaneously, the pretraining paradigm is making strides in the domain of multimodal data, enhancing performance across a diverse range of tasks, from cross-modal to fully multimodal. The upcoming section unveils recent architectures crafted to exploit this rich multimodal data.


\vspace{-0.4em}
\subsubsection {Transformer-Based Architectures}
Recent work on large scale pretraining has inspired the research community to progress in multimodal tasks \cite{radford2021learning}. The combined pretraining-based advanced approaches on image-text pairs at a large scale are developing rapidly which outperform task-specific architectures. For instance, in ViLBERT \cite{lu2019vilbert}, the authors pretrained BERT on textual and visual inputs in different streams that maintain relationships through co-attention layers. After pretraining on the Conceptual Captions dataset \cite{sharma2018conceptual}, the model is transferred to mainstream language-vision tasks, including VQA, VCR, and caption-based image retrieval and achieved consistent improvement over task-specific architectures. 
The Vision-Language Pretraining (VLP) in a unified manner is proposed by \cite{zhou2020unified} that outperformed the previous SOTA for image captioning and VQA tasks. The proposed model follows the shared encoder-decoder architecture  and is pretrained on a large scale image-text pairs in an unsupervised manner. They evaluated the model on VQA 2.0, COCO Captions, and Flickr30k \cite{flckr30k} Captions. In line with prior work , Li et al. \cite{li2020oscar} proposed a pretraining approach in which the object tags are used as anchors to enhance the alignment learning with paired text.

\begin{table*}[t!]
\centering
\caption{Transformer-based pretraining approaches, benchmarks, tasks and architectural settings. }
\label{tab:my-table-transformer}
\setlength\tabcolsep{1.5pt}
\begin{tabular*}{\linewidth}
{@{\extracolsep{\fill}}p{0.14\linewidth}p{0.26\linewidth}p{0.21\linewidth}p{0.33\linewidth}}
\toprule
\textbf{Model}    & \textbf{Benchmarks} & \textbf{Pretraining Tasks} & \textbf{Architectural Comparison} \\
\midrule
VLP \cite{zhou2020unified} & VQA2.0, COCO Captions, Flickr30k  & IC, VQA & 12-layer, 768-hidden, 12-heads, 110M param. \\ \hline

 VirTex \cite{desai2021virtex} & COCO Captions & Image Classification, IC & Visual backbone: ResNet-50; textual head: two unidirectional Transformers \\ \hline

ViLBERT \cite{lu2019vilbert}   & VQA, VCR, RefCOCO, IR  & VQA, VCR, GRE, CIR  & Textual Encoder: BERT-base; Visual backbone: Faster R-CNN (ResNet 101) 
  \\ \hline

ERNIE-ViL \cite{yu2020ernie}   & VCR, RefCOCO+, VQA, IR-Flickr30K, QR-Flickr30K & SGP, VCR , VQA, GRE, IR, TR  & Textual Encoder: BERT-base; Visual backbone: Faster R-CNN (ResNet 101)
  \\ \hline
  
OSCAR \cite{li2020oscar}   & COCO, CC, SBU captions, flicker30k, GQA  & VQA, ITR, IC, NOC, GQA, NLVR2  & Base: 12-layer, 768-hidd, 12-heads, 110M param.; Large: 24-layer, 1024-hidd, 16-heads, 340M param.
  \\ \hline

Vokenizer \cite{tan2020vokenization}   &  SST-2. QNLI, QQP, MNLI,SQuAD v1.1, SQuAD v2.0, SWAG Avg. & GLUE, SQuAD, SWAG, GLUE & Textual Encoder: BERT; Vision Encoder: ResNeXt-101-32x8d
  \\ \hline

AV-HuBERT \cite{shi2022learning} &  
LRS3, VoxCeleb2 & ASR & Hybrid ResNet-Transformer architecture \\ \hline

CLIP \cite{radford2021learning} & COCO, Visual Genome, YFCC100M & Image Classification, IC, OCR, AR, Geo-Localization & Textual Encoder: GPT-2, GPT-3; Vision Encoder: ViT-B/32, ViT-B/16, ViT-L/14
 \\ \hline

BLIP \cite{Li2022BLIP} & No-Caps, COCO, Flickr30k & ITR, IC, VQA, VD, NLVR & Textual Encoder: BERT base; Image Encoder: ViT-B/16, ViT-L/16
 \\ \hline

BLIP-2 \cite{li2023blip} &  No-Caps, COCO, Flickr30k, VQAv2, OK-VQA, GQA &  Instructed Zero-shot Image-to-Text Generation, VQA, IC, ITR & 
Textual Encoder: Q-former; Vision Encoder: FlanT5, ViT-L/14, ViT-g/14
 \\ 
\bottomrule
\vspace{-2em}
\end{tabular*}
\end{table*}

In addition to aforementioned  approaches, structured knowledge is extracted from scene graphs in the  recent research \cite{yu2020ernie}, which assist in learning joint representations as semantically connected features of text and vision. Scene graph prediction tasks which includes Object Prediction (OP), Attribute Prediction (AP), and Relationship Prediction (RP) are newly proposed by the authors in the pretraining phase. The proposed model can learn semantically aligned joint representation. The model's efficacy is evaluated across five cross-modal tasks, wherein it leads the VCR task, exhibiting a performance enhancement of 3.7\%. Desai et al. \cite{desai2021virtex} proposed VirTex, a data-efficient alternative to supervised pretraining that uses captions as a supervisory signal for vision tasks. They conducted joint pretraining of the model which comprises of convolution and transformer backbones from scratch for vision and language modalities which aims to generate captions for images. 
The visual backbone comprised of ResNet-50 is used to compute visual features of input images, which are then passed to the textual head to generate captions for corresponding images. After training, the textual head is discarded, as the primary aim is to utilize the learned features for subsequent visual tasks. Moreover, Shi et al. \cite{shi2022learning}  proposed Audio-Visual Hidden Unit BERT (AV-HuBERT), trained from scratch on LRS3 benchmark which consists of speech and visual features. The hybrid architecture ResNet-transformer serves as the backbone of the proposed model. AV-HuBERT extracts the phonetic and linguistic features from audio and visual streams simultaneously to produce latent representation. This representation can be used for pretraining baselines, e.g., HuBERT \cite{hsu2021hubert} which can outperform models that are pretrained on a single modality. Table  \ref{tab:my-table-transformer} draws the comparative picture of discussed approaches, on the basis of benchmarks, tasks and architectural settings respectively.

Word2Vec \cite{mikolov2013distributed}, GPT \cite{radford2018improving}, ELMO \cite{gardner2018allennlp}, BERT, and other SSL based frameworks perform good natural language understanding but does not consider grounding information from visual world as motivated by Bender et al. \cite{bender2020climbing} and Bisk et al. \cite{bisk2020experience}. The pretraining process designed by Tan et al. \cite{tan2020vokenization}, known as Vokenization, considers both language tokens and their corresponding visual information, termed as "vokens", as input for supervision. The author utilized the small image captioning dataset to train a vokenizer that can further generate language vokens of large corpora. The results achieved by visually supervised models outperformed SSL-based pretraining approaches on GLUE, SWAG, and SQuAD benchmarks. However, 
models trained using supervised methods often rely heavily on labeled data, which can limit their generality and usability. Therefore, to overcome these challenges, self-supervised multimodal learning methods have emerged, such as CLIP \cite{radford2021learning}. CLIP (Contrastive Language-Image Pre-training) is pretrained on WIT (WebImageText) \cite{radford2021learning}, which contains image-text pairs collected from the web. Unlike standard vision models, CLIP jointly trains an image encoder and a text encoder during the training phase. The performance of CLIP is benchmarked across over 30 vision and vision-language datasets, including COCO \cite{MS-COCO}, Visual Genome \cite{krishna2017visual}, and YFCC100M \cite{thomee2016yfcc100m} etc. BLIP (Bootstrapping Language-Image Pre-training) \cite{Li2022BLIP}, a unified vision-language framework, uses knowledge distillation on captions for improved performance. It achieves state-of-the-art results on various tasks, including image-text retrieval, image captioning, VQA, and even in zero-shot performance on text-to-video retrieval and VideoQA. BLIP is extensively evaluated on No-Caps \cite{agrawal2019nocaps}, COCO \cite{MS-COCO}, and Flickr30k \cite{flckr30k}. BLIP-2 \cite{li2023blip}, an improved visual-language model based on comprehensive pretraining strategy, achieves state-of-the-art performance across numerous vision-language tasks. BLIP-2 exhibits a wide range of zero-shot image-to-text abilities, which include visual knowledge reasoning, visual common sense reasoning, visual conversation, and personalized image-to-text generation. Like its predecessor, BLIP-2 uses the same evaluation datasets and continues to implement the Captioning and Filtering (CapFilt) method.

\vspace{-0.4em}
\subsubsection{Unifying Architectures}
Unifying architecture is designed to accept different modalities as input and train the model on multiple tasks to lessen the task-specific parameters as generic architecture. 
Li \textit{et al.} \cite{li2021towards} presented the unified model for text-only and vision-only tasks. The author put effort into creating a single suitable transformer-based pretrained model that can be finetuned on any modality for any downstream task.
The foundation model is a transformer pretrained jointly on unpaired images and text. The pretraining is based on knowledge distillation from the teacher pretrained model for better joint training and gradient masking for balancing the parameters’ updates. The shared transformer that can encode all modalities for the different tasks by employing a task-specific classifier is used. The aspiration is to reduce the task and modality-specific parameters by bringing up a more generic model, and maximum computation occurs in the shared transformer module.

The UNITER \cite{chen2020uniter} is a pretraining approach conducted at a large scale over four benchmark datasets. Joint embedding from both modalities is learned to perform heterogeneous downstream tasks. MLM, MRM (Masked Region Modeling), ITM (Image-Text Matching), and WRA (Word-Region Alignment) are employed as a pretraining task. Additionally, the pretraining task achieves global image text alignment using Conditional masking. Optimal Transport is the author's second concept for the WRA task to improve the alignment between images and words. Hu et al. \cite{hu2021unit} proposed a unified transformer for multitasking based on multimodal learning at a time. The proposed architecture uses a specialized encoder for each modality and a shared decoder for every task. DETR \cite{carion2020end} is used for visual features encoding and BERT \cite{devlin2018bert} performs the textual feature encoding. Contrastive learning is employed on multimodal data by Akbari et al. \cite{akbari2021vatt} to train a transformer encoder that processes audio, text, and video simultaneously.  

Wang et al. \cite{wang2022unifying} proposed the One For All (OFA) method, which unifies tasks and modalities via a sequence-to-sequence framework based on Unified vocabulary (for all kinds of the modality). OFA represents data of different modalities in a unified space that discretizes images and text to form a unified output vocabulary. They presented three qualities that a unifying model should support to maintain multitasking for any modality: (1) Task-Agnostic: handcraft instruction-based learning is utilized to achieve this property. (2) Modal-Agnostic:  single Transformer-based architecture uses globally shared multimodal vocabulary to make it modal agnostic. (3) Task comprehensiveness: pretraining conducted on various unimodal and multimodal tasks to achieve task comprehensiveness. The transformer is the backbone of the encoder-decoder unified network for the pretraining, finetuning, and zero-shot tasks. It considers multimodality and multitasking to make it more generalized for unseen tasks. OFA achieved SOTA on multimodal and outperforms other well-known pretrained for unimodal. As in GPT \cite{radford2018improving} and BART \cite{lewis2019bart}, BPE (Byte-Pair Encoding) is used to divide a sequence of words into sub-word sequences and embed them into features.

InstructBLIP \cite{dai2023instructblip} a general-purpose vision-language model, has achieved state-of-the-art, zero-shot performance on numerous vision-language tasks. InstructBLIP is a vision-language instruction tuning framework consisting of an image encoder, a Large Language Model (LLM), and a Q-Former \cite{li2023blip}. In its evaluation, InstructBLIP employs two LLMs: FlanT5 \cite{chung2022scaling} and Vicuna \cite{vicuna2023}. The evaluation metric incorporates 11 tasks across 28 datasets, including image captioning, video reasoning, visual conversational QA, knowledge grounded image question answering, video question answering, image captioning reading comprehension, image question generation, image classification, and LLaVa-Instruct-150k, a benchmark that incorporates visual conversation, complex reasoning, and detailed image description.

\begin{table*}[t!]
\centering
\caption{The unifying architectures with the benchmarks, tasks and architectural setting adopted by the authors.}
\label{tab:my-table-unifying}
\begin{tabular*}{\linewidth}{@{\extracolsep{\fill}}p{0.14\linewidth}p{0.27\linewidth}p{0.15\linewidth}p{0.3\linewidth}}
\toprule
\textbf{Model}    & \textbf{Benchmarks} & \textbf{Tasks} & \textbf{Architectural Comparison} \\
\midrule
UniT \cite{hu2021unit} & COCO, QQP, VQAv2 QNLI, MNLI, SST-2, SNKI-VE & VQA, OD, VE & 201M param., comprised of ResNet-50 + BERT\\
\hline
ViT-BERT \cite{li2021towards} & MNLI, QQP, QNLI, SST-2, RTE  Cifar, ImageNet, Flowers, Pet & Vision-only, text-only  &  12 layer transformer, 768-hidden, 3072 MLP \\
\hline
UNITER \cite{chen2020uniter}   & VQA, Flicker30k, NLVR, Ref-COCO & IR, TR VQA, RE & 12-24 layers, 768- 1024 hidden 12-16heads, 86-303M param.
\\\hline
VATT \cite{akbari2021vatt}    &  
AudioSet, HowTo UCF101, HMDB51, MSR-VTT 
& VAR, AEC, TVR     & Transformer-based variants; 155M to 415M param.\\ \hline

OFA \cite{wang2022unifying}      & SST-2, RTE, MRPC, QQP, QNLI, MNLI, SNLI-VE   & NLU, NLG,  Image Classification & Transformer-based variants; 33M to 940M param.\\ 

\hline

InstructBLIP \cite{li2023blip} &  No-Caps, Flickr30k, GQA,VSR, IconQA, TextVQA, Visdial, HM, VizWiz, SciQA IMG, MSVD QA, MSRVTT QA, iVQA
& IC, VR, ICRC, Video QA, IQG, 
Image QA,LLaVa-Instruct-150k & 
Variations of BLIP-2, Image encoder:ViT-g/14,LLMs:FlanT5-XL (3B), FlanT5-XXL (11B), Vicuna-7B and Vicuna-13B
 \\ 

\bottomrule
\vspace{-2em}
\end{tabular*}
\end{table*}
\vspace{-1em}
\subsection{Discussion}\label{sec3.3}
Most multimodal approaches uses a transformer as a backbone.
 These architectures have outperformed CNNs on most vision tasks as self-attention focuses on global contextual learning at a low-level stage. However, the generalization of large-scale multimodal architectures heavily depends upon the pretraining objective, learning technique, and nature of the benchmarks illustrated in this survey.\par
VLP \cite{zhou2020unified} is considered as the first SOTA on joint vision-language multi-modal tasks specifically generation and understanding. Likewise, VirTex showed SOTA performance on multimodal tasks \cite{desai2021virtex}. ViLBERT \cite{lu2019vilbert} marked SOTA on VQA and RefCOCO+ with huge margin. AV-HuBERT \cite{shi2022learning} used audio-visual representation on the benchmark (for audio-only speech recognition) and led to 40\% relative WER reduction over the state-of-the-art performance. ERNIE-ViL \cite{yu2020ernie} achieved SOTA on 5 cross-modal downstream tasks and ranked at first on the VCR leaderboard with an absolute improvement of 3.7\%. OSCAR \cite{li2020oscar} created new SOTA on six well-established vision-language understanding and generation tasks. Vokenizer \cite{tan2020vokenization} presented visually-supervised language models with consistent improvements on multiple language tasks. Finetuning on a similar benchmark offers outstanding results, though this needs more parameters, time, and resources. Many researchers are addressing these challenges by proposing unifying architectures to deal with multiple modalities with the same backbone for multitasking. Such generic models as summarized in Table \ref{tab:my-table-unifying} can share the parameters for different downstream tasks and avoid extensive pretraining or finetuning for each task. Although unifying architectures are not specialized for a specific task, they show competitive results. \par
It is observed that pretraining in a correlated manner considering different modalities enhance performance if strong alignment and correspondence among representation are considered during the training phase. The models trained on multiple modalities also express good performance on unimodal tasks as ViT-BERT \cite{li2021towards} surpassed the ViT on vision-only tasks. However, ViT-BERT could not surpass BERT on text-only tasks due to the low amount of data for finetuning, specifically on RTE dataset. Likewise, the results achieved by joint training of baselines give better performance than cross-modal finetuning, which faces the challenge of mismatching between pretraining objectives and downstream tasks. ViT-BERT achieved 83\% and 89\% Average scores on text-only and Vision-only tasks, respectively, making it a moderate option with an 86\% avg for both tasks (consume fewer parameters than UniT). The UniT \cite{hu2021unit} as a generic model achieved promising results with shared parameters on all tasks. However, UniT could not surpass VisualBERT or BERT, which are trained for specific tasks only. Smaller batch size and higher learning rate cause the problem of lower performance and divergence, respectively. \par
The UNITER-large \cite{chen2020uniter} model achieved comparable performance across all the benchmarks but consumed 303M parameters. UNITER-base model also performed well except VQA with 86M parameters. VATT \cite{akbari2021vatt}  outperformed the CNN-based approaches in all metrics for audio event recognition and video action recognition tasks. Additionally, it exhibits competitive results for text-to-video retrieval. The author introduced the DropToken strategy that reduces the computational complexity of training. DropToken is the sampling technique alternative to dimension and resolution reduction to mitigate redundancy challenges.
Additionally, Multimodal Contrastive Learning based on Noise Contrastive Estimation and Multiple Instance Learning is used to align video-audio and video-text pairs after projecting in the shared space. OFA \cite{wang2022unifying} surpassed the UNITER, UNIMO, and other SOTA on VQA and SNLI-VE. Notably, it achieved the highest score on RTE and RefCOC0+ by outperforming task-specific and unifying architectures. Specifically, OFA results assert that large-scale multimodal pretrained approaches compete with natural language pretrained SOTA for understanding and generation tasks. Furthermore, OFA outperformed Uni-Perceiver on out-of-domain tasks, e.g., single sentence and sentence pair classification.

InstructBLIP \cite{dai2023instructblip} proposed a novel instruction tuning framework towards generalized vision-language models. Their proposed method achieved state-of-the-art performance across an array of benchmarks with strong evaluation protocol. Besides, InstructBLIP has demonstrated its potential as an advanced initial model for finetuning on downstream tasks.

\vspace{-1em}
\subsection{Acronyms}\label{sec3.0}
This section contains all acronyms in the tables. 
\textbf{AEC}: Audio Event Classification, \textbf{Auton. Driving}: Autonomous Driving, \textbf{AR}:Action Recognition, \textbf{AVSS}: Audio Visual Speech Synthesis, \textbf{CIR}: Caption To Image Retrieval, \textbf{CMR}: Cross-Modal Retrieval, \textbf{ED}: Event Detection, \textbf{EL}: Entity Labeling, \textbf{EQA}: Embodied Question Answering, \textbf{GR}: Gesture Recognition, \textbf{GRE}: Generation of referring expressions, \textbf{IC}: Image Captioning, \textbf{ICRC}: Image Captioning Reading Comprehension,  \textbf{IE}: Information Extraction, \textbf{IQG}: Image Question Generation, \textbf{IR}: Image Retrieval, \textbf{IS}: Indoor Segmentation, \textbf{ITR}: Image to Text Retrieval, \textbf{LaVa-Instruct-150k}: includes visual conversation, complex reasoning, and detailed image description, \textbf{MML}: Multimodal Learning, \textbf{MMSA}: Multimodal Sentiment Analysis, \textbf{NLG}: Natural Language Generation, \textbf{NLU}: Natural Language Understanding, \textbf{NLVR}: Natural Language for Visual Reasoning, \textbf{OCR}: Optical Character Recognition, \textbf{OD}: Object Detection, \textbf{RE}: Referring Expression, \textbf{SSS}: Sound Source Separation, \textbf{TD}: Text Detection, \textbf{TR}: Text Retrieval, \textbf{TVR}: Text-To-Video Retrieval, \textbf{VAC}: Video Action Recognition, \textbf{VAR}: Visual to Audio Retrieval, \textbf{VC}: Video Captioning, \textbf{VCR}: Visual Common Sense Reasonng, \textbf{VD}: Visual Dialog, \textbf{VE}: Visual Entailment, \textbf{VG}: Visual Generation, \textbf{VLP}: Vision Language Pretraining, \textbf{VQA}: Visual Question Answering, \textbf{VR}: Video Reasoning, \textbf{VU}: Visual Understanding.

\section{Multimodal Applications}\label{sec4}
\begin{figure*}[t]
    \centering
  \includegraphics[width=14cm,height=7.2cm]{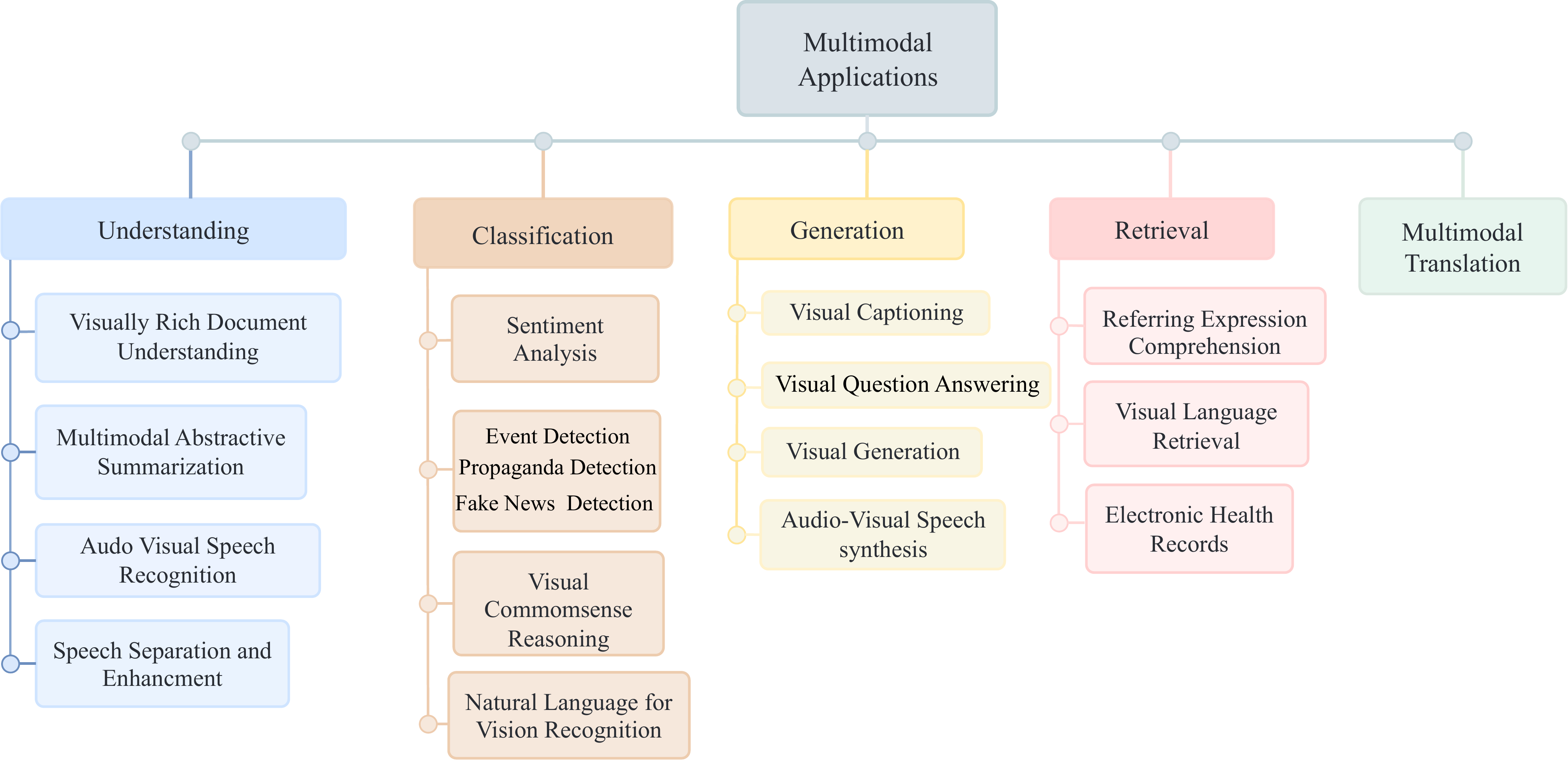}
  \caption{Taxonomy of the multimodal applications in Section \ref{sec4}.}
  \label{fig:4}
\end{figure*}

This section presents the categorical detail of multimodal applications enhanced by deep learning architectures as shown in Figure \ref{fig:4}. The multimodal tasks are divided into main categories: understanding, classification, retrieval, generation, and translation. 
The benchmark, evaluation metrics, description, and comparison for the best-performing architectures are discussed for each multimodal application.


\vspace{-0.5em}
\subsection{Understanding}
Understanding text, speech, and vision is the most important and primary task of AI-based systems. Relevant information extraction, recognition, and identification of entities from different mediums (text, images, speech) are its subtasks. The information achieved through understanding benefits many downstream tasks directly or indirectly. 
\vspace{-0.5em}
\subsubsection{Visually Rich Document Understanding}

Understanding visually rich documents with structured information is important for different applications \cite{im2021self}. Contrary to the classical information extraction methods, Visually Rich Documents Understanding considers the visual layout along with the text. LayoutLMv2 \cite{xu2020layoutlmv2} pretrained a Transformer that learns from different modalities at the pretraining stage by integrating and aligning layout, textual, and visual information. LayoutMv2 used more than 10 billion documents to pretrain a model that outperformed SOTA on document understanding tasks. More recently, Li \textit{et al.} \cite{li2021structext} proposed StrucTexT, which outperformed LayoutLMv2 by pretraining a Transformer in a self-supervised manner, with less parameters, achieved high scores. The publicly available benchmark datasets for downstream tasks are SROIE \cite{huang2019icdar2019} and FUNSD \cite{jaume2019funsd}. LayoutLMv2 achieved 97\% and 84\% F1 score on SROIE and FUNSD, respectively. StrucTexT had 96.88\% and 85.68\% F1 on SROIE and FUNSD, respectively.\par
Gu et al. \cite{gu2022xylayoutlm} released an improved version XYLayoutLM of LayoutLMv2 \cite{xu2020layoutlmv2}. They proposed Augmented XY Cut to capture reading orders as layout information, which was neglected previously. Furthermore, the variable length of the input sequence is dealt with DCPE (Dilated Conditional Position Encoding - inspired by CPE \cite{chu2021we}) that creates the 1D and 2D features of textual and visual input, respectively, for Convolution. The model succeeded in surpassing the other approaches on XFUN \cite{xu2020layoutlmv2} benchmark for SER (Semantic Entity Recognition) and RE (Relation Extraction) tasks.     
\vspace{-0.5em}
\subsubsection{Multimodal Abstractive Summarization}
This task takes massive multimodal content (video, images and corresponding text) from the internet and extracts the vital information to generate a summary \cite{liu2020multistage}. 
Palaskar et al. \cite{palaskar2019multimodal} generated a summary using Multisource seq2seq with Hierarchical attention that integrates information from different modalities. 
Likewise, a Multistage fusion network was proposed that established the interaction between different source modalities. 
Recently, an approach was proposed \cite{yu2021vision} that exploits the visual modality to generate a summary with Generative pretraining language models (GPLMs). 
The add-on layer based on attention is inserted in visually-guided GPLMs to maintain the visual incorporation and text generation.
The model was evaluated on the How2 \cite{sanabria2018how2} dataset, which contains instructional videos with two to three sentences of 2000 hours. The visually grounded variants of the proposed model outpaced the baselines, achieving the highest ROGUE-1, ROGUE-2, and ROGUE-L on the benchmark.
\vspace{-0.5em}
\subsubsection{Audio-Visual Speech Recognition}
Audio-visual Speech recognition (AVSR) is one of the earliest multimodal research domains that encouraged the research community to understand speech by using hearing and vision features simultaneously. Visual features and speech play a vital role in understanding speech, especially noise and assisting patients suffering from speech impairment. Most of the recent models wav2vec2.0 \cite{baevski2020wav2vec}, HuBERT \cite{hsu2021hubert} and De-CoAR2.0 \cite{ling2020decoar} were using audio only for speech recognition. Lip movement provides a supervisory signal by exploiting self-supervision to recognize better.\\
Recently released, Audio-Visual Hidden Unit BERT (AV-HuBERT) \cite{shi2022learning} is a self-supervised representation learning framework. The mentioned approach Learn audio-visual speech representation by taking favor of lip-reading and ASR. The proposed model achieved 32.5\% WER on LRS3 \cite{afouras2018lrs3} benchmark by using 30 hours of labeled data from 433 hours and achieving 26.9\% outperformed the former SOTA that trained on 31K hours \cite{makino2019recurrent}. The author also stated that exact representation performed speech recognition on audio-only tasks by reducing 40\% WER. The model is exceptional for visual only modality by exploiting contextualized representation of AV-HuBERT. 

\vspace{-0.5em}
\subsubsection{Speech separation and Enhancement}
Speech enhancement (SE) is the process in which speech signals of the target speaker are extracted in an acoustically noisy environment. SE improves speech quality (sounds) and speech intelligibility (linguistic content). The estimation of multiple targets in the speech is known as speech separation, or source separation \cite{prajwal2020learning}. 
Previously, these tasks were handled with statistical and mathematical criteria under signal processing \cite{rehr2017importance}. Currently, the evolution of supervised learning, multimodal methods, and fusion techniques attain the uninfluenced visual features of the speaker accompanying acoustic features in a noisy environment. 

Audi-Visual Sentence Extraction (AV-SE) and Audi-Visual Sentence Separation (AV-SS) systems count several speakers and trace their faces. Detection and Tracking algorithms are used to achieve  high-dimensional visual frames of  faces. The dimensionality is reduced with an active appearance model that is based on principal component analysis (PCA) \cite{cootes2001active}. 
Sound Source separation is achieved by Zhou et al. \cite{zhu2021leveraging} using the Appearance Attention Module that leverages categorical information on of single frame video. The author optimized the model by correlating the appearance embedding and feature maps. 
The sound source is located by the scalar product of embedding and feature maps. 
The dataset MUSIC contains YouTube videos made up of different musical instruments and has off-screen noise \cite{zhao2018sound}. 
The model is evaluated on the MUSIC dataset using standardized evaluation matrices, i.e., Signal to Distortion Ratio (SDR), Signal to Interference Ration (SIR), and Signal to Artifact Ratio (SAR). The attention-based approach surpassed the Resnet-18 and Resnet-50 by achieving SDR 10.74, SIR 17.29 and SAR 13.04. 

\vspace{-0.5em}
\subsection{Classification}
Classification is the systematic arrangement of samples into a group or category that follows the established criteria. Several tasks of vision, textual and speech fall under the umbrella of classification. The prominent multimodal classification tasks discussed in this study are sentiment analysis, fake news detection and event detection. 
\vspace{-0.5em}
\subsubsection{Sentiment Analysis}
Sentiment analysis is the core task in NLP that extract and classify reviews, feeling, gestures and behavior toward a specific entity. Sentiment Analysis understands people's perspectives for efficient decision-making in multiple domains. Textual representation has been analysed widely in previous research such as \cite{kim2016topic}, \cite{camacho2017role}, and \cite{wood2021market}. However, the sentiment analysis task moved from text modality to another form of modality due to social media and the internet. Chen \textit{et al.} \cite{chen2020swafn} designed deep learning architecture to extract sentiment from multimodal complex data by designing two component-based methods consisting of shallow fusion and aggregation parts. The shallow fusion component extracts contextual information from the different domains using the attention mechanism, and the aggregation part attains sentimental word-aware fusion. The proposed architecture outperformed other methods by achieving the highest score on multimodal datasets CMU-MOSI, CMU-MOSEI, and YouTube datasets. 

\vspace{-0.5em}
\subsubsection{Event Detection}
 Detection of an event, trend or situation specifically through content available on social media becomes more efficient and generous by considering data from different modalities \cite{hu2017adaptive}. A massive amount of data is generated that significantly impacts the lives, property, and psychology of humans \cite{algiriyage2022multi}. Event detection can be mapped to other realistic scenarios, such as Emergency Management, Disaster Detection, and Topic Detection \cite{xiao2022survey}. Even though the data from different perspectives enhance the performance, the challenges are also increased for methods to deal with redundant and heterogeneous characteristics.       
No large enough dataset is available for disaster detection to employ deep learning architecture except CrisisMMD \cite{alam2018crisismmd}, CrisisNLP and CrisisLex.
For traffic event detection, Chen \textit{et al.} \cite{chen2021multi} created a multimodal dataset by integrating the traffic-related filtered tweets with sensor data. The author achieved 84\%, 83\%, 87\% F1 score with CNN, RNN, and mmGAN (multimodal GAN) models, respectively.  
\vspace{-0.5em}
\subsubsection{Detecting Propaganda in Memes}
Propaganda is the type of communication that affects the psychology of people that leads them to keep specific opinions about any entity or perform some action. Due to the high usage of social media, it has become a societal and political problem. Memes that contain visual and textual content are being used as a significant fraction of the medium on the internet to trigger this issue. 

Dimitrov et al. \cite{dimitrov2021detecting} approached this problem as a multimodal and multi-label task by detecting the techniques to promote propaganda in memes. They released a new dataset consisting of 950 memes (containing both; textual and visual content) annotated with 22 different propaganda techniques. Previous Datasets addressed the propaganda at document level \cite{barron2019proppy}, sentence level and fragment level \cite{da2019fine}. SOTA models performed the experiments 
Experiments were conducted on variations of models that are pretrained on two approaches: (i) unimodally pretrained and (ii) pretrained with the multimodal objective. 
BERT and ResNet-152 are trained separately in the first variation and fused using MMBT (Multimodal bitransformers) with early, middle, or late fusion. 
ViLBERT and Visual BERT are pretrained in multimodal nature on Conceptual Captions and MS COCO, respectively. The results achieved by ViLBERT and Visual BERT surpassed the baseline and other variations. Results analysis expressed that transformer-based multimodal approaches are practical and produce efficient results. 
\vspace{-0.5em}
\subsubsection{Fake News Detection}
Fake news can use multimedia to cause panic on social media. The text and the image content misinterpret the facts and manipulate human psychology, leading to rapid propagation of fake news \cite{jin2017multimodal}. A multimodal architecture can detect fake news by looking for mismatches between the modalities \cite{zhou2020safe}.

The problem of detecting fake news for unseen and emerging events is addressed in \cite{wang2018eann} by leveraging an Adversarial Neural Network that uses an event discriminator to remove event-specific features and preserve shared features among events and modalities. Khattar et al. \cite{khattar2019mvae} proposed SpotFake, which utilizes BERT to learn textual features and VGG-19 (pre-trained on ImageNet) to learn visual features. They employ a simple concatenation technique to combine features obtained from different modalities. Similarly, \cite{khattar2019mvae} introduces the Multimodal Variational Autoencoder (MVAE), consisting of an encoder, a decoder, and a fake news detector module. The variational autoencoder utilizes probabilistic latent variable models by optimizing a bound on the marginal likelihood of the observed data. By leveraging the acquired multimodal representations from the bimodal variational autoencoder, the fake news detector classifies multimodal posts as genuine or fake.

Recently, Wang \textit{et al.} \cite{wang2022fmfn} proposed fine-grained multimodal fusion networks (FMFN) to detect fake news. First, CNN extracted visual features, and RoBERTa \cite{liu2019roberta} is used to get contextualized embedding of words. Then an attention mechanism is used between visual and textual features to enhance the correlation for fusing features. Finally, a binary classifier is adopted to perform detection on fused features.  
Their model achived 88\% accuracy on the Weibo dataset \cite{jin2017multimodal} by focusing only on tweets that contain text and images.  
\subsubsection{Visual Commonsense Reasoning} 
Visual Commonsense Reasoning performs robust Visual understanding by Integrating cognition, grounding, and language reasoning with recognition tasks. Zellers et al. \cite{zellers2019recognition} introduced the VCR dataset containing 290k MCQA (Multiple Choice Question Answer) problems. The author proposed the Recognition-to-Cognition Networks that consider the grounding and contextual information for cognition level visual understanding. The model surpassed the previous SOTA developed on VQA \cite{antol2015vqa}. Song et al. \cite{song2021kvl} proposed the Knowledge enhanced Visual and linguistic BERT that utilized the external commonsense Knowledge. This approach outperformed the BERT-based approaches and previous VCR models by marginal improvement. 
\subsubsection{Natural Language for Vision Recognition}
The NLVR (Natural Language Vision Recognition) task checks the relationship consistency between the provided one text description and multiple images. Previous pretraining approaches \cite{tan2019lxmert, chen2020uniter, huang2020pixel} considered NLVR as binary classifier. Vision Language Navigation is a similar task in which agents traverse the real-world dynamics by following linguistic instructions. Zhu et al. Proposed AuxRN (Auxiliary Reasoning Navigation) \cite{zhu2020vision} based on self-supervision that Observe Semantic Information from surrounding for Vision-Language Navigation. In other words, the model learns the implicit information from the environment, estimates the navigation, and predicts the next position. 

\vspace{-0.5em}
\subsection{Generation}
In multimodal generation tasks, the data provided by the input modality is processed to generate another modality or paraphrased into the same modality as the output. The tasks concealed by generative category heavily rely on the previously mentioned understanding and classifications tasks. The image captioning task generates the text after classifying the object in the input image. Likewise, the text-to-image generation task understands the input text to propose a target image. 
\vspace{-0.3em}
\subsubsection{Visual Captioning} 
Visual captioning is the task of converting an image or video modality into a related text modality. The input is image pixels, and the output is syntactically and semantically meaningful text. Producing captions based on images bridges the gap between low and high semantic features found in image and text modalities. Some SOTA datasets in image captioning are MS-COCO \cite{MS-COCO} and Flickr30k \cite{flckr30k}. The baseline and conventional visual captioning pipeline consist of an image encoder and a text decoder.
Recent work leverages BERT capabilities, such as in \cite{bert_caption}. BERT models \cite{devlin2018bert} are initially trained using a large textual corpus, then fused textual and visual modalities embedding are finetuned using the right-masking pretraining technique. Vinyals et al. \cite{vinyals2015show} presented an end-to-end system based on a neural network and combined SOFA sub-networks for vision and language models. The method in this article significantly exceeds the performance of previous methods. In 2017, Chen et al. \cite{chen2017sca} improved CNN and proposed SCA-CNN based on images that CNN can extract: Spatial, Channe-wise, and Muli-layer. SCA-CNN can incorporate Spatial and Channel-wise attention in a CNN and dynamically modulate the sentence generation context in multi-layer feature maps. Unlike previous studies, Rennie et al.  \cite{rennie2017self} used reinforcement learning to optimize image captioning systems. They proposed that optimizing the CIDEr using metricSCST(self-critical sequence training) is highly effective. 
\vspace{-0.3em}
\subsubsection{Visual Question Answering}
Visual Question Answering (VQA) attempts to answer linguistic questions by retrieving information from visual cues \cite{VQA}. VQA combines information from written questions and high-dimensional visual images or videos. Questions could vary from simple true/false to knowledge-based and open-ended questions, whereas visuals could vary from a simple sketch or image to a video. Furthermore, VQA combines functions from NLP and CV fields such as language understanding, relation extracting, attribute and object classifying, counting, knowledge-base, and commonsense reasoning \cite{recent_paper}.  
Baseline VQA maps question text and visual embeddings obtained via recurrent and convolutional neural networks (RNN and CNN), respectively, to a common vector space. Mapping embedded representation to a shared vector space enables VQA to tackle open-ended free-form questions. In the literature, VQA deep learning techniques are classified as joint embedding models \cite{vqa_emb}, attention mechanism \cite{vqa_att}, compositional model \cite{subramanian2019analyzing}, Graph Neural Networks (GNN) \cite{gao2020multi} and knowledge base model\cite{vqa_kb}.     
There exist review papers targeting VQA as a research field of its own such as in \cite{VQA} and \cite{VQA2}. In addition to surveys, benchmarks such as \cite{bench1} by Carnegie Mellon University (CMU) and \cite{bench2} are available. Benchmark datasets are VAQ v1.0 \cite{VQA2}, VAQ-X \cite{vqa-x}, and VAQ-CP \cite{vqa-cp}. 
\vspace{-0.3em}
\subsubsection{Visual Generation}
Visual Generation, also known as text-to-image generation, typically uses the input text to generate images. Visual generation is the reverse direction of image captioning. In 2016, Reed et al. \cite{reed2016generative} first proposed deep convolutional generative adversarial networks(GAN) to synthesize images based on text descriptions. The training model is based on
DC-GAN, but different from traditional GAN, the input of D is added with real image and false text description pairs. The author trained a CNN to predict style using an image generated by generator G. The predicted style can be used in the composition of G. The dataset used in this paper are: Caltech-UCSD Birds\cite{wah2011caltech} dataset, Oxford-102 Flowers dataset, and MS COCO \cite{MS-COCO} dataset. 
Xu et al. \cite{xu2018attngan} introduced Attentional Generative Adversarial Network (AttnGAN) for fine-grained text-to-image generation. AttnGAN, composed of an attentional generative network and Deep Attentional Multimodal Similarity Model (DAMSM), focuses on relevant words to depict specific image subregions. DAMSM augments generator training with a fine-grained image-to-text matching loss. Comprehensive evaluation shows AttnGAN significantly outperforms previous GAN models. This method is evaluated on CUB\cite{wah2011caltech} and COCO \cite{MS-COCO} datasets.
\vspace{-0.3em}
\subsubsection{Audio-Visual Speech Synthesis}
Prajwal et al. \cite{prajwal2020learning} presented the model synthesizing speech from lip movement in the presence of noise. The author released the benchmark Lip2Wav Dataset, as the previous work \cite{qu2019lipsound} was evaluated on small and limited vocabulary-based datasets \cite{afouras2018lrs3, afouras2018deep,harte2015tcd}. The designed dataset enables the model to synthesize speech from unconstrained lip movements.

High-quality speech is generated \cite{ping2017deep} using a Tacotron \cite{shen2018natural} inspired decoder that produce melspectrogram from text inputs. It is conditioned on the face embeddings encoded in the previous representation, and outperformed previous work \cite{ephrat2017vid2speech, akbari2018lip2audspec, vougioukas2019video} on all objective metrics for lip-to-speech work especially on TIMIT \cite{harte2015tcd} dataset. 


\vspace{-0.4em}
\subsection{Retrieval} Retrieval systems take the input from the user as a query and provide the most relevant results by considering the contextual information. The evolution of multimedia-enhanced is 
in the form of well-established cross-modal tasks.
\vspace{-0.6em}
\subsubsection{Referring Expression Comprehension}
In Referring Expression Comprehension (REC) application, an expression is used to refer to and localize a target object in an image. The referring expressions are specific and detail the object properties and the relationship to its surroundings. Therefore, visual attributes, relationships, and contextual information need to be addressed. In \cite{9285213}, REC supervised DL architectures are categorized into one or two stages. In two-stage pipelines, the first stage generates and lists proposed objects based on the image. The second stage encodes the referring expression and computes a matching score between the proposed objects and encoded referring expressions. In one-stage pipelines, image and language features are concatenated and fed into the model in a single faster step. Some pretrained models for task-agnostic vision and language applications used for REC are VL-BERT \cite{Su2020VL-BERT} and ViLBERT \cite{lu2019vilbert}.
\vspace{-0.6em}
\subsubsection{Visual Language Retrieval}
Indexing, query formulation, retrieval and evaluation are the steps required to build an Information Retrieval (IR) application.  In indexing and query formulation, documents and user interfaced queries are represented by their characteristic features, respectively \cite{EHR}. The retrieval system then maps both representations to retrieve or extract the required useful information. The performance of the retrieval task is evaluated based on recall and precision. In multimodal information retrieval, the system searches documents with different modalities such as text, images, videos, or physiological signals and images. Employing more than one modality enriches information retrieving processes. One example of multimodal information retrieval is in electronic health records. 
\subsubsection{Electronic Health Records}
Patients health records contain various modalities such as categorical data, text, images such MRI scans, or signals such as electrocardiogram (ECG) \cite{EHR}. Information retrieval system can extract information from different modalities to report, present, and/ or predict a patient health status. For instance, Supervised Deep Patient Representation Learning Framework (SDPRL) engages different modalities information to learn patient representation \cite{zhang2021learning}. SDPRL is build and tested using the benchmark  dataset MIMIC-III. Chen et al.  \cite{chen2022ms} address the problem of Major Depressive Disorder (MDD) detection by proposing a novel Graph Neural Network (GNN)-based multimodal fusion strategy, coined as "modal-shared modal-specific GNN". This approach accounts for heterogeneity/homogeneity among psychophysiological modalities and inter/intramodal characteristics, and it employs a reconstruction network and attention mechanism to obtain a compact multimodal representation. Similarly, Multimodal Graph Neural Network  framework was introduced to predict cancer survival using multimodal data such as gene expression, copy number alteration, and clinical data \cite{gao2021predicting}.
\vspace{-0.4em}
\subsection{Multimodal Translation (MMT)}
Visual modality has been considered marginally beneficial for machine translation due to the absence of sufficient features in the image.
Caglan et al. \cite{caglayan2019probing} proposed that visual information benefits Machine Translation (MT) when the source sentence lacks linguistic context. 
The author exploited the Degradation technique in the form of Color Deprivation, Entity Masking and Progressive Masking to degrade the source sentence. 
Additionally, unrelated images (violation of semantic compatibility) were fed to evaluate the visual sensitivity of the approach. ResNet-50 CNN based encoder \cite{he2016deep} is used for visual features extraction. Multimodal Attention for Neural Machine Translation (NMT) inspired by \cite{bahdanau2014neural} achieved the context vector from textual and visual features. 
Multi30K and flicker30 were used for training. 
MMT performed better than NMT. The authors presented a novel encoder for Neural Machine Translation (NMT) \cite{gao2020multi} that employs a graph-based approach for combining multiple modalities. This approach takes advantage of detailed semantic connections between these different data types to enhance the learning of combined representations.

Su et al. \cite{su2021multi} introduced an approach that exploits the interaction of semantic representations across modalities, drawing inspiration from QA and NER domains. Two attention-based models were proposed: one using a bi-directional attention mechanism to learn from text and visual feature interaction, and the other incorporating a co-attention mechanism that generates context vectors from textual representation and subsequently uses it to update the text context vector with further interactive information. These models were evaluated on the expanded Multi30K dataset, showing superior performance over previous baselines, with or without pretraining.

\vspace{-0.3em}
\section{Dataset}\label{sec5}
This section summarizes the benchmarks for the pretraining, finetuning, and evaluation of Multimodal models. As per our knowledge, we cover all the benchmarks containing image, text, video, and audio modalities. The motive is to provide ready-to-use information to the new researchers regarding the nature of benchmarks, the number of samples in each benchmark, the usability for different tasks, and the comparative information as shown in Table \ref{tab:my-table}. The MS COCO dataset \cite{MS-COCO} is used for image recognition, image detection, image segmentation, and image captioning. For each image in the dataset, five different descriptions are provided. This dataset contains 91 object types that would be easily recognizable. The dataset includes more than 300,000 images and 2.5 million labeled instances.

The Flickr30k \cite{flckr30k} is collected from Flickr, together with 5 image descriptions provided by human annotators. The Flickr8k \cite{hodosh2013framing}is smallest version, also collected from Flickr, that leads to train model easier and faster. FUNSD \cite{jaume2019funsd} is a small dataset comprising 199 real, fully annotated, scanned forms. FUNSD contains 31485 words, 9707 semantic entities, and 5304 relations and aims to extract and structure forms' textual content. SROIE \cite{huang2019icdar2019} consists of a dataset with 1000 whole scanned receipt images and annotations for the competition on scanned receipts OCR and key information extraction (SROIE). CMU Multimodal Opinion Sentiment and Emotion Intensity (CMU-MOSEI) dataset \cite{zadeh2018multimodal} is the largest dataset of multimodal sentiment analysis and emotion recognition to date. In 2018, the CMU-MOSEI dataset was created, consisting of 23,453 annotated video clips. These clips were sourced from 1,000 unique YouTube videos that covered 250 different topics. 

VQA (Visual Question Answering) \cite{antol2015vqa} is a large dataset containing more than 250,000 images with at least 3 questions per image and 10 ground-truth answers per question. VQA v1 provides 6,141,630 ground-truth answers and 1,842,489 plausible answers, making it harder for the model to answer the questions correctly. VAQ-cp (Visual Question Answering under Changing Priors) is the splits of the VQA v1 and VQA v2 datasets. CrisisMMD \cite{alam2018crisismmd} is a sizeable multimodal dataset of natural disasters collected from Twitter, including earthquakes, hurricanes, wildfires, and floods that happened in the year 2017 across different parts of the World. It has three types of annotations. The first one is "informative" or "Not informative", which determines whether it is helpful for humanitarian aid. The second type is "Humanitarian Categories", such as infrastructure and utility damage, vehicle damage, etc. The last type is "Damage Severity Assessment", which describes the severity of the damage.

MIMIC (Medical Information Mart for Intensive Care) \cite{johnson2016mimic} is a publicly available dataset developed by the Laboratory for Computational Physiology that comprises de-identified health data associated with thousands of intensive care unit admissions. Currently, it has three versions: MIMIC-II, MIMIC-III, and MIMIC-IV. MIMIC-III contains information about 53,423 adults admitted to critical care units from 2001 to 2012, such as the patient's gender, height, and other essential information, such as blood routine, liver function, and other hospital test data, as well as medication information.

\begin{table*}[h!]
\centering
\caption{The Benchmarks' Information for the tasks in this study. The Task column lists the tasks in the corresponding papers.}
\label{tab:my-table}
\begin{tabular*}{\linewidth}{lllll}
\toprule
\textbf{Dataset} & \textbf{Modality} & \textbf{Count}  & \textbf{Tasks}    & \textbf{Data Category}\\ 
\midrule
MS COCO \cite{MS-COCO} & text + image    & 328k    & \begin{tabular}[t]{@{}l@{}} Image Recog., IC \end{tabular}    & Multiple    \\

Conceptual Captions \cite{sharma2018conceptual} & text + image & 3.3M   &   VQA, VCR, IR, IC & Multiple \\

Flickr30K {\cite{flckr30k}}  & text + image    & 31k    & \begin{tabular}[t]{@{}l@{}} IR, CMR, IC  \end{tabular}    & Multiple    \\

Flickr8K \cite{hodosh2013framing} & text + image & 8k & \begin{tabular}[t]{@{}l@{}} IR, CMR, IC \end{tabular}  & Multiple \\ 
 
FUNSD \cite{jaume2019funsd} & text + image & 199 &  \begin{tabular}[t]{@{}l@{}}TD, OCR, EL \end{tabular} & Scanned forms \\

SROIE \cite{huang2019icdar2019} & text + image & 1k &  \begin{tabular}[t]{@{}l@{}} OCR, IE\end{tabular} &  \begin{tabular}[t]{@{}l@{}}Scanned images\end{tabular}                   \\ 
 
CMU-MOSEI \cite{zadeh2018multimodal}  & text + video & 23,453 &  
\begin{tabular}[t]{@{}l@{}}MMSA \end{tabular} & YouTube videos  \\ 

VQA \cite{antol2015vqa} & text + image & 250k   & VQA   & Multiple     \\ 

CrisisMMD \cite{alam2018crisismmd} & text + image & 16,097  & 
\begin{tabular}[t]{@{}l@{}}IC, ED\\ \end{tabular} & Disasters tweets \\ 

MIMIC \cite{johnson2016mimic} & \begin{tabular}[t]{@{}l@{}}MML data \end{tabular} & 53,423 & IR  & Patients Information \\ 

Fashion-200K \cite{fashion200k} & text + image & 200k & \begin{tabular}[t]{@{}l@{}}IC, IR \end{tabular} & Clothes \\ 

NYU Depth v1 \cite{silberman11indoor}, v2 \cite{Silberman:ECCV12} & RGB + Depth  & 4GB, 90GB  & \begin{tabular}[t]{@{}l@{}}IS, IC \end{tabular}  & Indoor Scenes \\ 

SKIG \cite{wada2020pre}  & text + image & 2160  & GR & Hand gestures \\ 

GoodNews \cite{biten2019good}  & text + image  & 466,000 & IC    & News  \\ 

MSR-VTT \cite{xu2016msr} & text + video & 200k & VU  & Commercial videos \\ 

MSVD-QA \cite{xu2017video} & text + video & 2089 & \begin{tabular}[t]{@{}l@{}}VR, VQA, VC \end{tabular} & Multiple \\ 

TGIF-QA \cite{jang2017tgif}   & text + image (gif) & 103,919 & VQA & Multiple \\ 

EQA-v1 \cite{yu2019multi} & text + 3D env. & 9,000    & \begin{tabular}[t]{@{}l@{}} EQA, VQA\end{tabular}  & Multiple  \\ 

VideoNavQA \cite{cangea2019videonavqa} & text + 3D env. & 28 & \begin{tabular}[t]{@{}l@{}} EQA, VQA\end{tabular} & Multiple \\

TDIUC \cite{kafle2017analysis} & text + image & 1,654,167 & VQA  & Multiple \\

nuScenes \cite{nuscenes}  & image + sensor data & 1.4M                        & \begin{tabular}[t]{@{}l@{}} IR, Auton. Driving\end{tabular}  & Driving Information \\ 

CUB-200 \cite{wah2011caltech} & text + image  & 11,788 & VG   & Birds \\ 

Oxford-102 Flowers \cite{Nilsback08} & text + image  & 4,080-26,316         & VG & Flowers \\ 

VCR \cite{zellers2019recognition} & text + image   & 263k  & VU & Moive Scenes \\ 

How2 \cite{sanabria2018how2}  & audio + text + video  & 2000 hours & MML &  Instructional Videos     \\ 

Lip2Wav \cite{prajwal2020learning} & audio + video  & 120 hours &  AVSS &  Talking Face Videos\\

MUSIC \cite{zhao2018sound} & audio + video  & 685   &   SSS & Videos of Music\\

MOSI \cite{zadeh2016mosi} & audio + text + video  & 3702   &   Multimodal analysis & YouTube videos\\

GRID \cite{zhang2020can} & audio + video  & 33,000   &   VSR & Speech\\

\bottomrule
\end{tabular*}
\vspace{0em}
\end{table*}

Fashion200K \cite{fashion200k} contains 200K fashion images, and each image comes with a compact attribute-like product description. MIT-Stata Center \cite{fallon2013stata} is a multimodal dataset containing vision (stereo and RGB-D), laser and proprioceptive data. This dataset comprises over 2.3 TB, 38 h and 42 km. This dataset also includes ground-truth position estimates of the robot at every instance. This is an instrumental dataset for robotic mapping and CV research. The NYU-Depth dataset consists of video sequences recorded by the RGB and Depth cameras from the Microsoft Kinect. NYU-Depth v1 \cite{silberman11indoor} provides about 4GB of labeled data and about 90GB of raw data. NYU-Depth v1 includes 64 different indoor scenes and 7 scene types. NYU-Depth v2 \cite{Silberman:ECCV12} includes 464 different indoor scenes and 26 scene types.

The Shefeld Kinect Gesture (SKIG) \cite{wada2020pre} is a gesture dataset containing 2160 hand gesture sequences. These hand gestures can be classified into 10 categories:  circle (clockwise), triangle (anti-clockwise), up-down, right-left, wave, "Z", cross, come-here, turn-around, and pat. The sequences are recorded under 3 different backgrounds and 2 illumination conditions, which provide diversity. GoodNews \cite{biten2019good} is a large dataset containing 466,000 images with captions, headlines, and text articles. However, different from datasets like MSCOCO or Flicker8k, GoodNews includes a single ground truth caption per image. GoodNews captions written by expert journals have a longer average length than generic captioning datasets, meaning that these captions are more descriptive.

MSR-VTT \cite{xu2016msr} is a large-scale video description dataset: 10K web video clips with 38.7 hours and 200K clip-sentence pairs. MSR-VTT was created from 257 popular queries from a commercial video search engine. Each clip in MSR-VTT is annotated with approximately 20 natural sentences. This dataset is presented for video understanding. The Microsoft Research Video Description Corpus (MSVD) dataset MSVD-QA contains 122K descriptions of 2089 short video clips (usually less than 10 seconds). The MSVD dataset contains different language descriptions, such as English, Hindi, Romanian, Slovene, etc. MSVD-QA is a benchmark for video retrieval, visual question answering, and video captioning.

TGIF-QA \cite{jang2017tgif} is a large-scale dataset containing 103,919 QA pairs collected from 56,720 animated GIFs. These GIFs are from the TGIF dataset. The TGIF dataset is based on GIFS data as GIFs have a concise format and cohesive storytelling. TGIF-QA can be used for visual question-answering research. EQA (Embodied Question Answering) v1.0 \cite{yu2019multi} is a dataset containing 9,000 questions from 774 environments. The visual questions and answers in this dataset are grounded in House3D. EQA-v1 contains location, color, and place preposition questions.

VideoNavQA \cite{cangea2019videonavqa} is also a dataset used to study the EQA task. VideoNavQA contains 28 questions belonging to 8 categories with 70 possible answers. The complexity of the questions in VideoNavQA far exceeds that of similar tasks that use generation methods that extract ground truth information from the video to generate questions. Task Directed Image Understanding Challenge (TDIUC) \cite{kafle2017analysis} is a dataset containing 167,437 images and 1,654,167 question-answer pairs. TDIUC divides VQA into 12 constituent tasks, which makes it easier to measure and compare the performance of VQA algorithms. The 12 different question types are grouped according to these tasks.

nuScenes \cite{nuscenes} is a large-scale public dataset for an autonomous driving dataset with 3d object annotations. It is also a multimodal dataset. nuScenes provides 1.4 million camera images, 1500h of driving data from 4 cities (Boston, Pittsburgh, Las Vegas and Singapore), sensor data released for 150h (5x LIDAR, 8x camera, IMU, GPS), detailed map information, 1.4M 3D bounding boxes manually annotated for 23 object classes, etc. nuScenes can be used for intelligent agent research. nuImages is a large-scale autonomous driving dataset with image-level 2d annotations. It has 93k video clips of 6s each, 93k annotated and 1.1M un-annotated images. The Caltech-UCSD Birds dataset (CUB-200) \cite{wah2011caltech} includes images of 200 different bird species, totaling 11,788 images. Every image in this dataset has annotations that include a bounding box around the bird, a basic segmentation of the bird, and labels for various attributes. The Oxford-102 Flowers dataset features images from 102 bird categories, with each category containing between 40 to 258 images. These images display a wide range of sizes, poses, and lighting conditions. 

Visual Commonsense Reasoning (VCR)\cite{zellers2019recognition} contains over 212K (training), 26K (validation), and 25K (testing) questions, answers, and rationales derived from 110K movie scenes. It is widely used for cognition-level visual understanding. How2\cite{sanabria2018how2} is a multimodal dataset containing instructional videos with English subtitles and crowdsourced Portuguese translations. How2 covers a wide variety of topics across 80,000 clips (about 2,000 hours). Lip2Wav\cite{prajwal2020learning} dataset contains 120 hours of talking face videos across 5 speakers. This dataset has about 20 hours of natural speech per speaker and vocabulary sizes of over 5000 words for each. MUSIC (Multimodal Sources of Instrument Combinations) dataset\cite{zhao2018sound} contains 685 untrimmed videos of musical solos and duets. The dataset spans 11 instrument categories: accordion, acoustic guitar, cello, clarinet, erhu, flute, saxophone, trumpet, tuba, violin and xylophone. MOSI\cite{zadeh2016mosi}(Multimodal Opinionlevel Sentiment Intensity) dataset contains 93 videos and 3702 video segments. The dataset provides not only sentiment annotations, but also manual gesture annotations. GRID\cite{zhang2020can} consists of video recordings from 54 speakers, with 100 utterances per talker, 33,000 utterances in total.

\section{Future Forecasting}\label{sec6}
This section describes the techniques researchers have observed from previous studies to adopt and consider for achieving better results and reducing the computational cost. Additionally, these approaches have not been considered or explored thoroughly in recent studies even after expressing efficient performance in relevant research.

\textbf{Building Unified Models.} Previous studies proved the effectiveness of transformers for NLP, CV, and multimodal tasks. Recently, researchers have presented a unified architecture with a single agent transformer serving as the backbone and deal with all modalities for multiple tasks \cite{baevski2022data2vec}. Even though the results are impressive on downstream tasks, still could not surpass task-specific models for some of the tasks \cite{singh2022flava}. For zero-shot learning, the performance highly depends on instructions. More optimized instructions can lead to more satisfactory results. Researchers are trying to address the challenge of sensitivity that the model expresses with the slight changes in prompts and parameters.

\textbf{Leveraging GNNs for multimodal pretraining.}
Ektefai et al. \cite{ektefaie2023multimodal} proposed Multimodal Graph Learning (MGL) to manage multimodal data input and produce a common output representation for a variety of downstream tasks. This methodology initially identifies relevant entities across data modalities, projecting them into a shared namespace, and subsequently combines these different modalities. 
The process applies a message-passing module to learn node representation based on intra-modality and inter-modality adjacency matrices. Finally, through an aggregation process, it generates node, subgraph, or graph-level representations suitable for downstream tasks.

\textbf{Pretraining Strategy.} Developing an optimal set of pretraining objectives is worth exploring for multimodal tasks. Pretext tasksdirectly influence the performance for downstream tasks. While selecting pretraining objectives, we need to consider modalities, datasets, network architecture, and downstream tasks. There is a potential to apply efficient stage-wise pretraining for multimodal methods as Bao et al.\cite{wang2021vlmo} successfully adopt it for a single modality. 

\textbf{Enhancing Pretraining with Multilingual Features.} Multilingual pretraining of multimodal systems is less examined as English-only multimodal benchmarks are experimented heavily. UC2 \cite{ zhou2021uc2}, M3P \cite{ ni2021m3p}, and MURAL \cite{ jain2021mural} exploit multilingual features by adding a separate encoder for multilingual or translated data. CCLM \cite{ zeng2022cross} based on ALBEF \cite{ li2021align} proposed an approach that outperformed the SOTA in a zero-shot cross-lingual set-up. In short, the multilingual perspective is essential for enhancing the scaling success of multimodal pretraining frameworks \cite{ chen2022pali}.

\textbf{Prompt Tuning.} Prompt tuning received attention after GPT-2 and GPT-3. This technique converts implicit information to a question or descriptive text, which assists in exploiting the textual probability of language models. Prompt learning with a pretraining objective (MLM) can address two challenges of finetuning i) – reduce the computational complexity by avoiding the different parameters for each downstream task ii) – reduce the gap between the representation learning of pretraining tasks and downstream tasks \cite{tsimpoukelli2021multimodal}. This technique is not explored for multimodal approaches yet. Technically, prompt tuning may transform the unifying architecture to a more normalized and generalized version.

\textbf{Masking Stage.} The interaction of different modalities faces alignment challenges. Masking is an effective strategy used to address alignment problem by generating text and images. Masking can be applied at different stages e.g., input level, task level but according to Zhuge et al. \cite{zhuge2021kaleido}, embedding level masking is the most effective stage for aligning the representation of different modalities. Fine-grain representation can be learned if we focus on the optimal alignment among audio, video and text modalities.

\textbf{Knowledge Infusion.} There is a need to fuse external knowledge in multimodal representation for visual, language and audio to make the model well-informed, as Chen et al. \cite{chen2023vlp} utilized illustrative knowledge for vision and language. More intelligent architectures are required that integrate knowledge base features for multimodal pretraining and downstream tasks. Knowledge Distillation enhanced the performance of ViT-BERT on all tasks by offering effective learning representation compared to other variants that do not exploit it. For SST-2 and IN-1K, accuracy improved 10\% and 5\%, respectively. Therefore, while designing pretraining objectives and cognitive architectures, commonsense, situational knowledge, hierarchical, structural, or network information must be considered.

\textbf{Considering Anticipated Evaluation.} Deep learning architectures need high computational resources for experiments. Current evaluation methods provide information about the effectiveness of the models after extensive experiments. For large-scale multimodal methods, there must be evaluation strategies to examine the model at an earlier stage \cite{long2022vision} to verify whether models are compatible with downstream tasks before paying the computation cost.  

\textbf{The acceleration and scaling of Multimodal architectures.} Aside from Knowledge distillation \cite{fang2021compressing}, pruning and quantization are still unexplored to compress and accelerate the model and improve the cross-modal inference speed. Inspired by the performance of sizeable pretrained language models, efforts are made to achieve success on multimodal tasks through extensive training and complex architectures. However, well-developed benchmarks and models are required to cope with the multimodal at a large scale with higher proficiency \cite{jia2021scaling, fei2021wenlan}.

\textbf{Including Audio Stream.} Recent Multimodal surveys \cite{long2022vision, du2022survey} specifically for pretraining have not considered the audio stream. Audio can be semantically rich like text and can provide emotional and supplementary information about the speaker, including the boundary data in the case of multiple speakers. Pretraining with audio is extremely important for unifying architectures as it makes the model capable of performing downstream tasks that contain the audio stream \cite{michelsanti2021overview}. However, the challenges of alignment and correspondence increase with multiple modalities. 
\vspace{-0.3em}
\section{Conclusion}\label{sec7}
This survey covered the role of deep multimodal learning and architectures to effectively deal with advanced multimodal tasks.
We started with deep learning-based task-specific multimodal architectures based on encoder-decoder, attention, and reinforcement learning. Then, we covered the advancements in hardware, large-scale computation resources, and pretraining approaches. Pretraining and finetuning alleviate various challenges of multimodal and cross-modal tasks, including multitasking. Therefore, we discussed the types and tasks of pretraining, which make the training procedure effective, and SOTA transformer-based pretraining approaches that produce high impact recently, summarize comparatively.
We further showed that the research community focused on pretraining at a large scale to create unified and generic architectures using a transformer as the backbone to perform a more complex task with identical parameters. We covered SOTA multimodal architectures for different multimodal applications and their performance on benchmarks. The applications showed the usefulness and the practicality of multimodal systems. Multimodal possibilities are enormous, yet so far, the capabilities are explored only for the English language. A possible suggested future research trajectory is to construct multimodal datasets in other languages and build multilingual frameworks.

\bibliographystyle{unsrt}
\bibliography{main}

\end{document}